\documentclass[10pt,twocolumn,letterpaper]{article}

\usepackage{cvpr}
\usepackage{times}
\usepackage{epsfig}
\usepackage{graphicx}
\usepackage{amsmath}
\usepackage{amssymb}

\usepackage{url}
\usepackage{graphicx}
\usepackage{amsmath}
\usepackage{amssymb}
\usepackage{url}
\usepackage[dvipsnames]{xcolor}
\definecolor{newcolor}{rgb}{.8,.349,.1}
\usepackage[fleqn]{mathtools}
\usepackage{enumerate}
\usepackage{enumitem}
\usepackage[export]{adjustbox}
\usepackage{framed}
\usepackage{subcaption}
\usepackage[export]{adjustbox}
\usepackage{xcolor}
\usepackage{array}
\usepackage{pifont}
\usepackage{soul}
\usepackage{xspace}
\usepackage{float}
\usepackage{placeins}
\definecolor{newcolor}{rgb}{.8,.349,.1}

\definecolor{airforceblue}{rgb}{0.36, 0.54, 0.66}
\definecolor{antiquebrass}{rgb}{0.8, 0.58, 0.46}
\definecolor{brown(traditional)}{rgb}{0.59, 0.29, 0.0}
\definecolor{calpolypomonagreen}{rgb}{0.12, 0.3, 0.17}
\definecolor{coolblack}{rgb}{0.0, 0.18, 0.39}

\usepackage[breaklinks=true,bookmarks=false]{hyperref}

\cvprfinalcopy 

\def\cvprPaperID{****} 

\begin{document}

\title{DILIE: Deep Internal Learning for Image Enhancement}

\author{Indra Deep Mastan and Shanmuganathan Raman\\
Indian Institute of Technology Gandhinagar\\
Gandhinagar, Gujarat, India\\
{\tt\small \{indra.mastan, shanmuga\}@iitgn.ac.in}}

\maketitle

\begin{abstract}
We consider the generic deep image enhancement problem where an input image is transformed into a perceptually better-looking image. Recent methods for image enhancement consider the problem by performing style transfer and image restoration. The methods mostly fall into two categories: training data-based and training data-independent (deep internal learning methods). We perform image enhancement in the deep internal learning framework. Our Deep Internal Learning for Image Enhancement framework enhances content features and style features and uses contextual content loss for preserving image context in the enhanced image.  We show results on both hazy and noisy image enhancement. To validate the results, we use structure similarity and perceptual error, which is efficient in measuring the unrealistic deformation present in the images. We show that the proposed framework outperforms the relevant state-of-the-art works for image enhancement.
\end{abstract}

\begin{figure*}[!h] 
\begin{center}
\begin{subfigure}[b]{0.15\textwidth} \includegraphics[width=\linewidth]{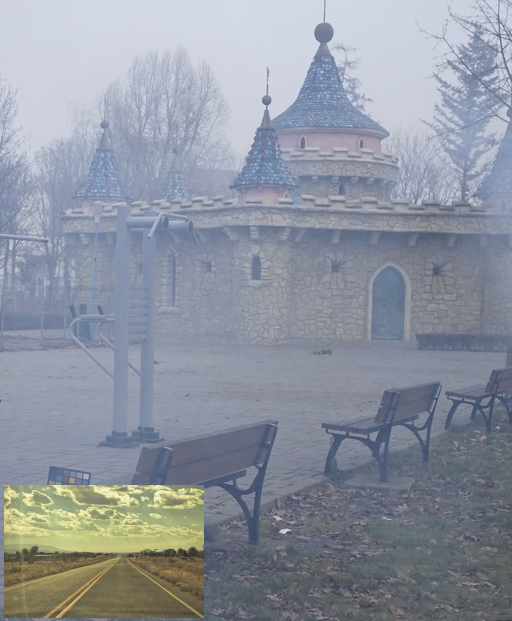}  \end{subfigure}
\begin{subfigure}[b]{0.15\textwidth} \includegraphics[width=\linewidth]{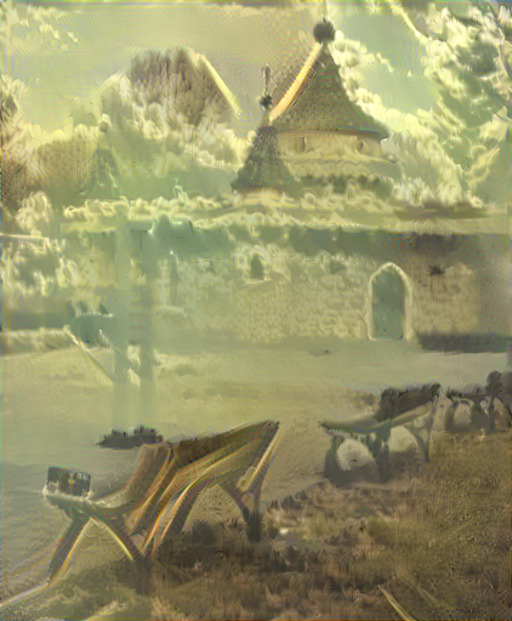} \end{subfigure}
\begin{subfigure}[b]{0.15\textwidth} \includegraphics[width=\linewidth]{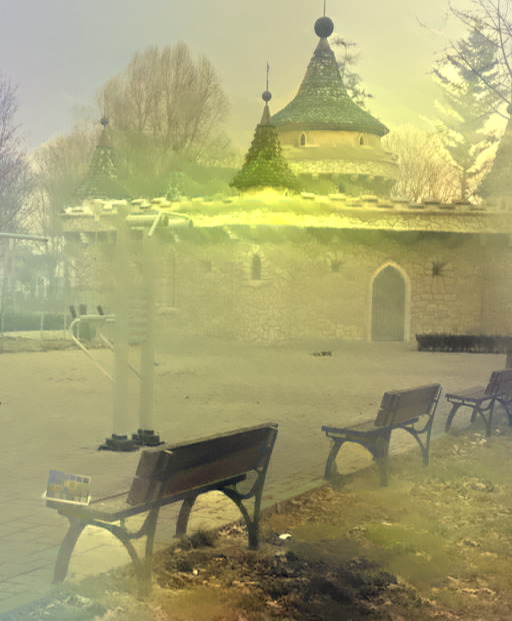} \end{subfigure}
\begin{subfigure}[b]{0.15\textwidth} \includegraphics[width=\linewidth]{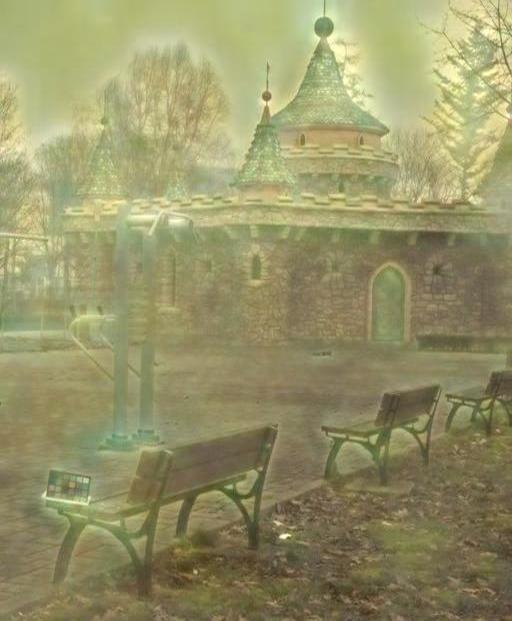} \end{subfigure}
\begin{subfigure}[b]{0.15\textwidth} \includegraphics[width=\linewidth]{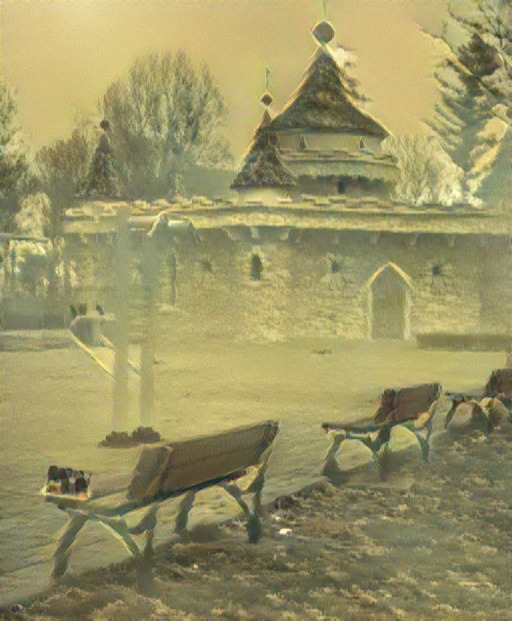}  \end{subfigure}
\begin{subfigure}[b]{0.15\textwidth} \includegraphics[width=\linewidth]{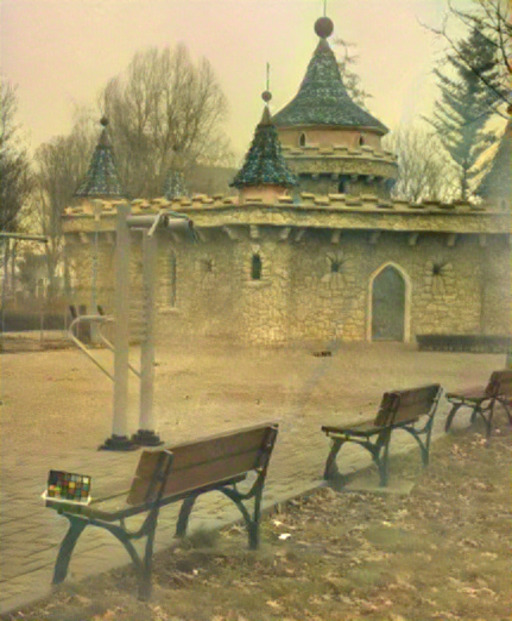} \end{subfigure} \\
\begin{subfigure}[b]{0.15\textwidth}
\includegraphics[width=\linewidth]{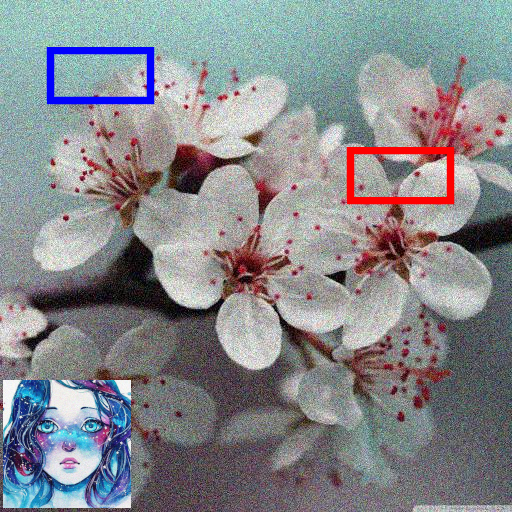} \\
\includegraphics[width=0.46\linewidth, cfbox=red 0.5pt 0.5pt]{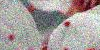}
\includegraphics[width=0.46\linewidth, cfbox=blue 0.5pt 0.5pt]{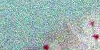}
\caption{\footnotesize  Content \& Style}  \end{subfigure}
\begin{subfigure}[b]{0.15\textwidth}
\includegraphics[width=\linewidth]{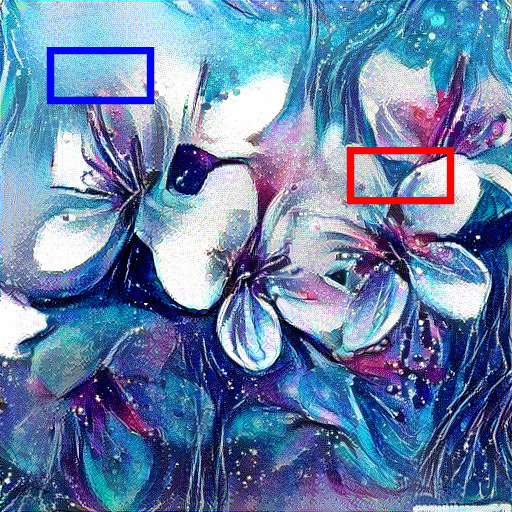} \\
\includegraphics[width=0.46\linewidth, cfbox=red 0.5pt 0.5pt]{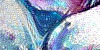}
\includegraphics[width=0.46\linewidth, cfbox=blue 0.5pt 0.5pt]{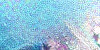}
\caption{\footnotesize  Neural style \cite{gatys2016image}}  \end{subfigure}
\begin{subfigure}[b]{0.15\textwidth}
\includegraphics[width=\linewidth]{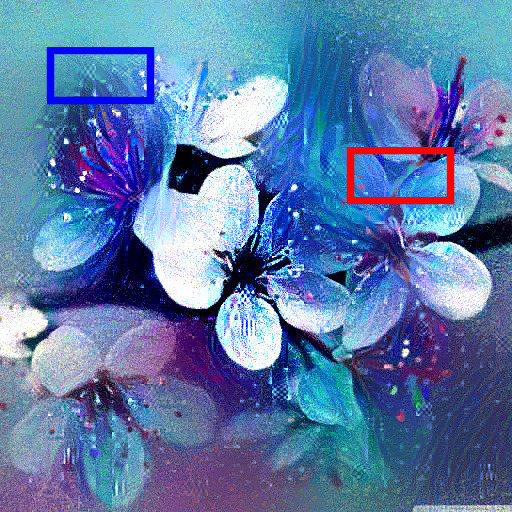} \\
\includegraphics[width=0.46\linewidth, cfbox=red 0.5pt 0.5pt]{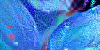}
\includegraphics[width=0.46\linewidth, cfbox=blue 0.5pt 0.5pt]{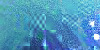}
\caption{\footnotesize  DPST \cite{luan2017deep}}  \end{subfigure}
\begin{subfigure}[b]{0.15\textwidth}
\includegraphics[width=\linewidth]{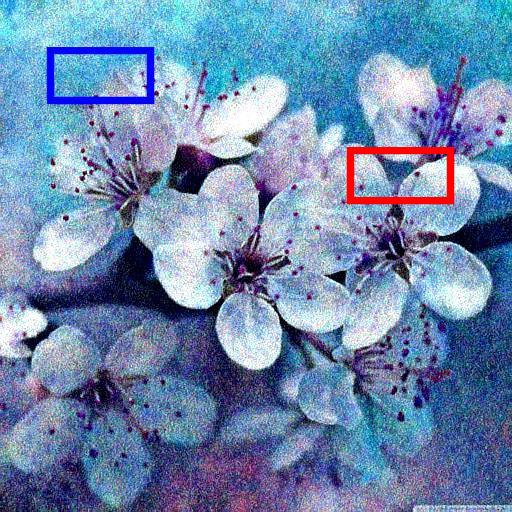} \\
\includegraphics[width=0.46\linewidth, cfbox=red 0.5pt 0.5pt]{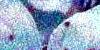}
\includegraphics[width=0.46\linewidth, cfbox=blue 0.5pt 0.5pt]{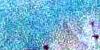}
\caption{\footnotesize  WCT2 \cite{yoo2019photorealistic}} \end{subfigure}
\begin{subfigure}[b]{0.15\textwidth}
\includegraphics[width=\linewidth]{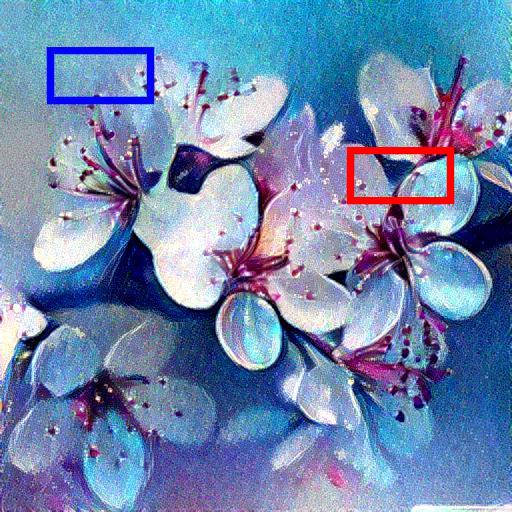} \\
\includegraphics[width=0.46\linewidth, cfbox=red 0.5pt 0.5pt]{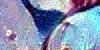}
\includegraphics[width=0.46\linewidth, cfbox=blue 0.5pt 0.5pt]{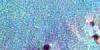}
\caption{\footnotesize  STROTSS \cite{kolkin2019style}}  \end{subfigure}
\begin{subfigure}[b]{0.15\textwidth}
\includegraphics[width=\linewidth]{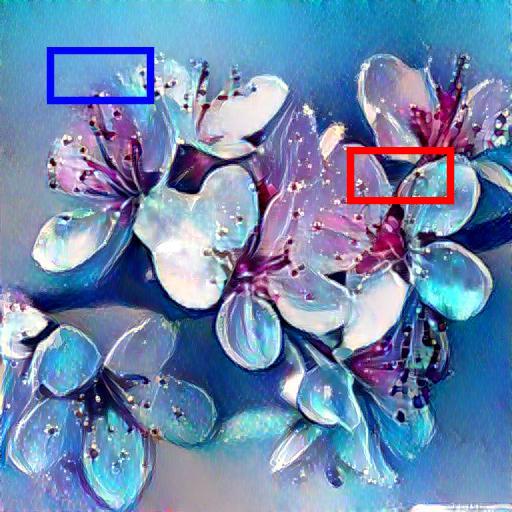} \\
\includegraphics[width=0.46\linewidth, cfbox=red 0.5pt 0.5pt]{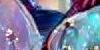}
\includegraphics[width=0.46\linewidth, cfbox=blue 0.5pt 0.5pt]{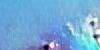}
\caption{\footnotesize  DILIE (ours)}  \end{subfigure} \vspace*{-0.15cm}
\caption{The figure shows that the deep internal learning-based DILIE framework output images with better perceptual quality. The style image is shown at the left corner of the content image in (a). The first row shows that DILIE output image with better perceptual quality for the enhancement of the hazy image. The second row shows DILIE output images with better clarity for noisy image enhancement.}\label{fig: DIELogo}
\end{center}
\end{figure*}

\section{Introduction}
Many computer vision tasks could be formulated as image enhancement tasks where the aim is to improve the perceptual quality of the image. For example, an image denoising method enhances image features and remove noise. The image style transfer method enhances the content image by transferring style features from a style image.

Deep image enhancement is an ill-posed problem that aims to improve the perceptual quality of an image using a deep neural network \cite{li2018lightennet, wang2020ebit, li2020underwater, yin2020novel}. An image could be considered as the composition of content features and style features. The content features denote the objects, their structure, and their relative positions. Style features represent the color and the texture information of the objects. Deep image enhancement aims to improve the quality of the content and the style features.

The image features may get corrupted in various ways. For example, bad weather conditions, camera shake, and noise, etc. Let us discuss an example of a deep image enhancement task. Consider a hazy image denoted by $I$. Haze particles degrade both the content features and the style features. The content features are corrupted because haze particles reduce the clarity of the structure of the objects. Style features are corrupted due to gray and blueish patterns introduced by haze. The enhancement of content and style features can draw inspiration from the image restoration and the style transfer methods. 

Image enhancement task is to improve the perceptual quality of hazy image $I$. The challenge is the haze particles are non-uniformly spread over the scene. One strategy is to utilize the content features from $I$ and transferring the photo-realistic features from a style image $S$. The interesting observation here is that maintaining the balance between the content feature and the style feature is challenging (Fig.~\ref{fig: ablation}).

Performing deep image enhancement without using paired samples of training data was proposed as an open problem \cite{zhang2017image}. Here, paired-samples indicate the instances of the original image and corrupted image pairs. Recent advancement in \emph{deep internal learning} (DIL) solves the open problem for image restoration and image synthesis tasks \cite{shocher2019ingan, ulyanov2018deep}. We categorize DIL methods for simplicity as: image reconstruction models \cite{ulyanov2018deep}, layer separation models \cite{gandelsman2019double}, and single image GAN frameworks  \cite{mastan2020dcil, shocher2019ingan}. The deep image enhancement of an image (content) is also performed using a style image in the style transfer \cite{gatys2016image, luan2017deep, kolkin2019style}.

We formulate a generic framework called Deep Internal Learning for Image Enhancement (DILIE). It does not use paired samples of training data and aims to learn features internally to perform image enhancement. Fig.~\ref{fig: DIELogo} shows the deep image enhancement performed for hazy and noisy images. The good perceptual quality of DILIE framework is due to the ability of CNN to learn good quality image statistics from a single image \cite{shocher2019ingan, ulyanov2018deep, gandelsman2019double}.  

\begin{figure*}[!h] \centering
\includegraphics[width=0.92\linewidth]{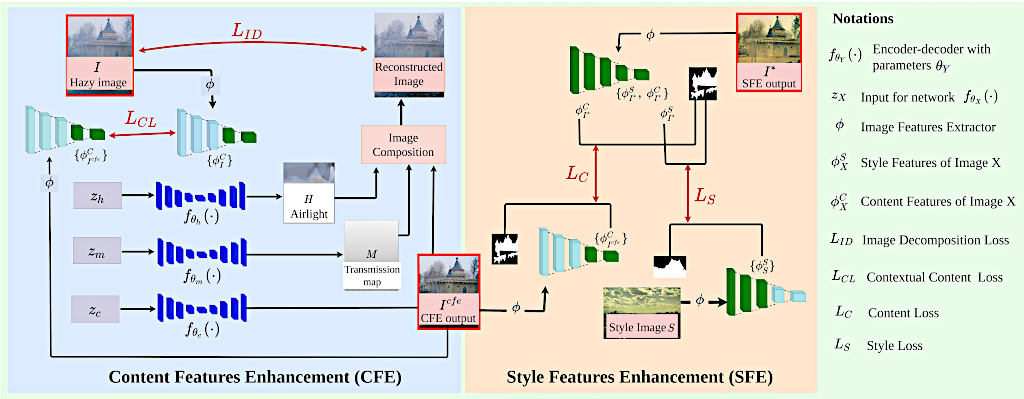}
\caption{\small \textbf{Image Dehazing.} The figure shows the pictorial representation of DILIE framework for the enhancement of the hazy image. The hazy image $I$ is transformed into an enhanced image $I^*$. The left side shows the content feature enhancement (CFE) and the right side shows the style feature enhancement (SFE). CFE performs image decomposition to output haze-free image $I^{cfe}$, transmission map $M$ and haze layer $H$. VGG19 network $\phi$ is used to extract features to compute contextual content loss $\mathcal{L}_{CL}$, content loss $\mathcal{L}_C$, and style loss $\mathcal{L}_S$. Image  decomposition loss $\mathcal{L}_{ID}$ is a pixel-based loss (Eq.~\ref{eq: decomposition}). $\mathcal{L}_{CL}$ uses contextual similarity criteria on content features (Eq.~\ref{eq: context}). SFE improves style features using content loss $\mathcal{L}_C$ and style loss $\mathcal{L}_S$. We describe DILIE framework in Sec.~\ref{sec: methodology}.}\label{fig: model}
\end{figure*}

We have illustrated DILIE framework in Fig.~\ref{fig: model} for hazy image enhancement. It takes the degraded image $I$ as input and generates the enhanced image $I^*$.  The \emph{main idea} is to formulate the content feature enhancement (CFE) and the style feature enhancement (SFE) models separately for generalizability. Fig.~\ref{fig: model} shows CFE decomposes the hazy image $I$ into environmental haze layer $H$ and haze-free image $I^{cfe}$. SFE transfers photo-realistic features from style image $S$ to $I^{cfe}$.  We describe the DILIE framework in Sec.~\ref{sec: methodology}. 

CFE is modeled based on the type of corruption. CFE, using the image decomposition model, is used for image dehazing and CFE using image reconstruction for image denoising. The image decomposition model performs joint optimization to separate the degraded image into clean and corrupted features. Image reconstruction generates a clean image with pixel-based reconstruction loss. Both these approaches rely upon the strong image prior captured by the encoder-decoder network.

The aim of SFE is to transform the input image (content) into a visually appealing output image by transferring style features from the style image. SFE is modeled based on the desired style specification, \textit{i.e.}, photo-realistic style transfer \cite{luan2017deep, yoo2019photorealistic} or artistic style transfer \cite{kolkin2019style}. Note that the distortions in the style transfer output lead to a lack of photo-realism. We measure the deformations using perceptual error Pieapp \cite{prashnani2018pieapp} computed between the content image and the output image. DILIE output images with low perceptual error (Table~\ref{table: PieAPP_NIMA}).


One of the important requirements for image enhancement is to preserve the context of the input image. DILIE preserves the semantics of the input image by comparing the context vectors. The context vectors represent high-level content information and are computed from feature extractor VGG19 \cite{mechrez2018contextual}. DILIE framework computes the contextual content loss $\mathcal{L}_{CL}$ to preserve the semantics of the scene. Fig.~\ref{fig: model} illustrates here $\mathcal{L}_{CL}$ is computed between $I$ and $I^{cfe}$ to preserve the contextual content features in $I^{cfe}$.

\begin{figure*}\centering 
\begin{subfigure}[b]{0.145\linewidth} \includegraphics[width=\linewidth]{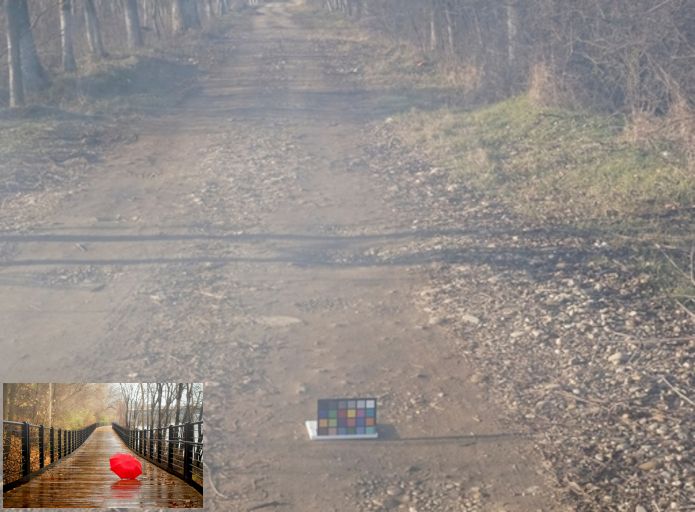} \caption{\footnotesize Content \& Style } \end{subfigure}
\begin{subfigure}[b]{0.145\linewidth} \includegraphics[width=\linewidth]{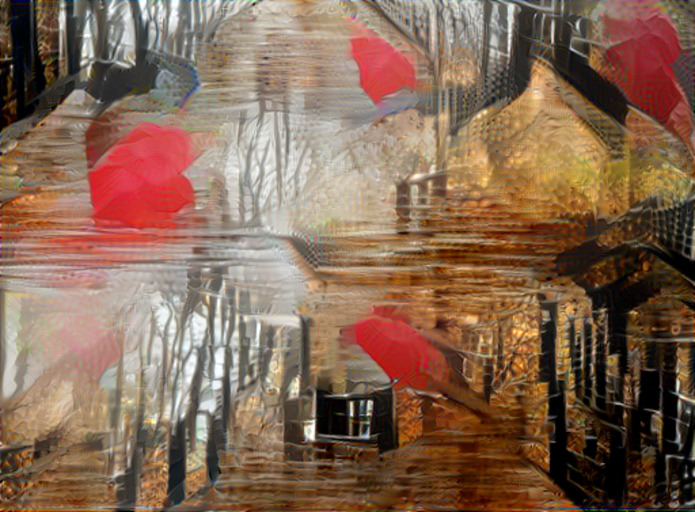}  \caption{\footnotesize  Neural style  \cite{gatys2016image}}\end{subfigure}
\begin{subfigure}[b]{0.145\linewidth} \includegraphics[width=\linewidth]{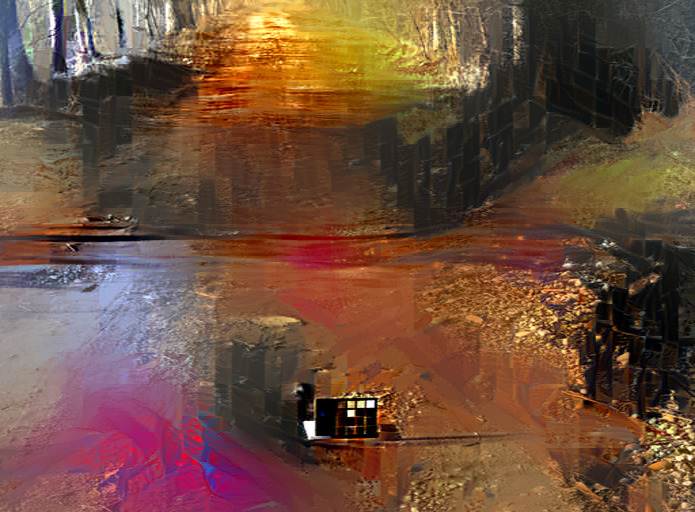} \caption{\footnotesize DPST \cite{luan2017deep} }\end{subfigure}
\begin{subfigure}[b]{0.145\linewidth} \includegraphics[width=\linewidth]{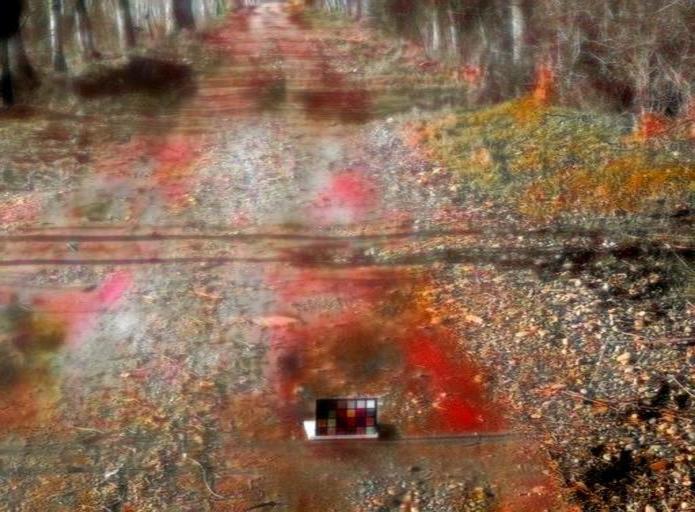}\caption{\footnotesize WCT2 \cite{yoo2019photorealistic} } \end{subfigure}
\begin{subfigure}[b]{0.145\linewidth} \includegraphics[width=\linewidth]{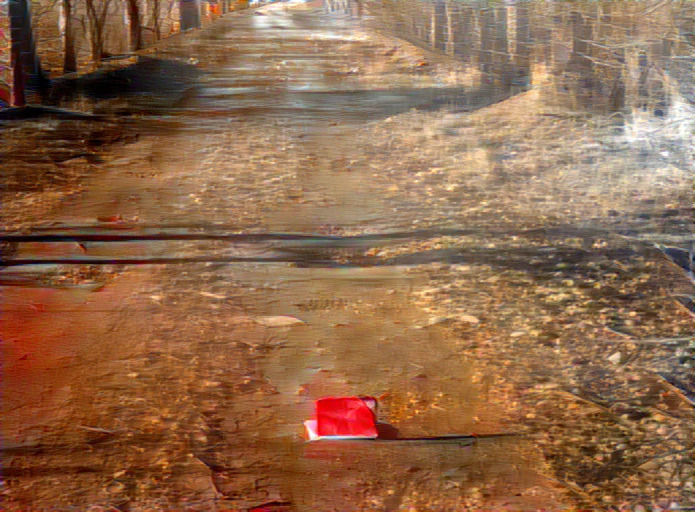} \caption{\footnotesize STROTSS \cite{kolkin2019style}}\end{subfigure}
\begin{subfigure}[b]{0.145\linewidth} \includegraphics[width=\linewidth]{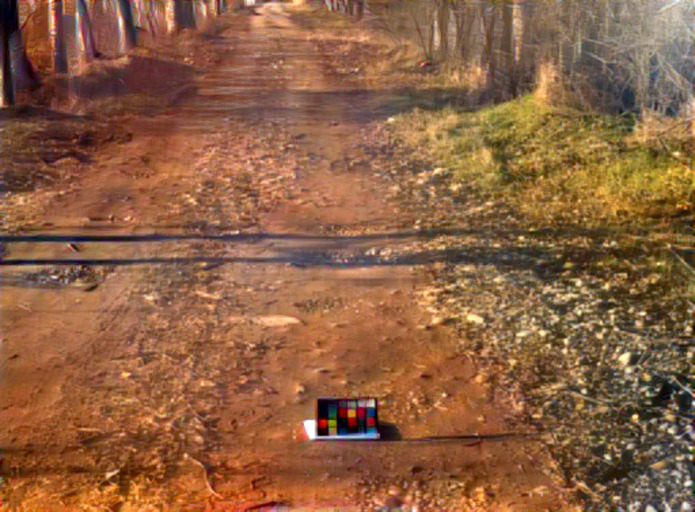}\caption{\footnotesize DILIE (ours)} \end{subfigure}\vspace*{-0.2cm}
\caption{\textbf{Hazy Image Enhancement (outdoor).} The content image contains haze and the style images are clear images (bottom left corner). The style images are photo-realistic. Neural style \cite{gatys2016image} deforms the geometry of the objects. DPST \cite{luan2017deep} does not distribute image features well. WCT2 \cite{yoo2019photorealistic} output contains haze corruption, as shown by white spots. STROTSS \cite{kolkin2019style} does not preserve fine image features details. It could be observed that DILIE (ours) output images with better visual quality {\color{blue} (the images are best viewed after zooming)}. }
\label{fig: ohaze}
\end{figure*}

We propose a generic deep internal learning framework (DILIE) that addresses corruption specific image enhancement using image reconstruction and image decomposition, and photo-realistic feature enhancement. We summarize the major contributions as follows.
\begin{itemize}[noitemsep,leftmargin=*]
\item We show that utilization of contextual features improves image dehazing and outperform relevant state-of-the-art works (Table~\ref{table: SSIM}). 
\item We show image enhancement for the challenging scenario where photos were taken in hazy weather (Fig.~\ref{fig: ohaze} and Fig.~\ref{fig: ihaze}). DILIE also performs enhancement of the noisy images (Fig.~\ref{fig: denoising}). 
\item DILIE outputs images with better visual quality and lower perceptual error (Table~\ref{table: PieAPP_NIMA} and Fig.~\ref{fig: ablation}).
\end{itemize}

\section{Related Work}
\noindent \textbf{Deep Internal Learning.} Recent DIL approaches perform image synthesis and image restoration without using paired samples for training \cite{ulyanov2018deep, shocher2019ingan}. The aim is to learn the internal patch distribution \cite{shocher2019ingan} and utilize the deep image prior \cite{ulyanov2018deep}. DIL is different from training data-based methods that use prior examples to supervise the image enhancement task \cite{johnson2016perceptual,  mechrez2018contextual}.

\noindent \textbf{Content Feature Enhancement.} CFE is performed using image reconstruction and image decomposition models. The structure of the encoder-decoder network (ED) provides application-specific image prior implicitly \cite{ulyanov2018deep}. Image reconstruction models use ED for denoising, super-resolution, and inpainting. Dehazing is formulated as an image decomposition problem \cite{gandelsman2019double}, where ED computes the image layer and haze layer separately. For simplicity, image dehazing methods could be classified into classical \cite{fattal2008single, he2010single}, supervised method using deep learning \cite{li2017aod}, and unsupervised methods \cite{gandelsman2019double}. 

\noindent \textbf{Style Feature Enhancement.} The related works for style feature enhancement are discussed as follows. Gatys et al. proposed Neural style \cite{gatys2016image} for style feature enhancemnt. Luan et al. \cite{luan2017deep} improved Neural style \cite{gatys2016image} for photo-realism. WCT2 enhances photorealism using wavelet transforms \cite{yoo2019photorealistic}. STROTSS \cite{kolkin2019style} uses optimal transport for more general style transfer.

\section{Our Approach} \label{sec: methodology}
DILIE is a unified framework to restore the content features and synthesizes new style features for the image enhancement task. Let us denote the input image by $I$. The DILIE framework is defined in Eq.~\ref{eq: DILIEFramework}.
\begin{equation}\label{eq: DILIEFramework}
I^* = \text{DILIE}(I, f, S, \phi, \alpha, \beta).
\end{equation}
Here, $I^*$ is the enhanced image. The encoder-decoder network $f$ is used for the reconstruction or decomposition of input $I$. The style image $S$ is used to enhance the style features of image $I$. The VGG19 network $\phi$ is used for image context learning \cite{mastan2020dcil} and the style features enhancement \cite{gatys2016image, kolkin2019style}. DILIE framework performs content feature enhancement (CFE) and style features enhancement (SFE) separately.  $\alpha$ and $\beta$ are the parameters used for CFE and SFE. CFE enhances content features by learning deep features using encoder-decoder $f$ (Sec.~\ref{ssec: CFE}). SFE uses a style image $S$ for photo-realistic and artistic feature enhancement (Sec.~\ref{ssec: SFE}).

Fig.~\ref{fig: model} shows DILIE framework for hazy image enhancement. CFE performs image decomposition to decompose hazy image $I$ into content feature $I^{cfe}$, haze layer $H$, and transmission map $M$. The image composition block combines the decomposed image features and outputs the reconstruction of $I$. It is done to preserve relationship between $I^{cfe}$, $H$, and $M$. SFE outputs the enhanced image as $I^*$. We discuss the components of the DILIE framework as follows.

\subsection{Content Feature Enhancement} \label{ssec: CFE}
CFE could be majorly performed in the following two ways: image reconstruction (IR) and image decomposition (ID). The formulation of content feature enhancement is given in Eq.~\ref{eq: model}. 
\begin{equation}\label{eq: model}
I^{cfe} = \text{CFE}(I, f, \phi, \alpha). 
\end{equation}
Here, $I^{cfe}$ denotes the output of the content feature enhancement. The structure of the encoder-decoder network $f$ provides an implicit image prior for the restoration of image features. The corruption specific image prior enables diverse applications, \textit{e.g.}, dark channel prior for the image dehazing \cite{gandelsman2019double, he2010single} and encoder-decoder without skip connections as denoising prior \cite{ulyanov2018deep}. The VGG network $\phi$ is used to extract the contextual features to compute the contextual content loss for preserving the context of image $I$ in CFE output. The parameter $\alpha$ denotes whether CFE is used to model image decomposition ($\alpha=1$) or reconstruction ($\alpha=2$).

\subsubsection{Image Decomposition} \label{ssec: ID}
Image decomposition (ID) improves the quality of images by separating image features and corrupted features. Formally, given an image $I$ as a combination of image feature layer and environmental noise. ID separate $I$ into the image features layer $I^{cfe}$ and the image corruption layer $D$, where the separation is determined by a mask $M$. In the image dehazing task, the mask is a transmission map that determines image features $I^{cfe}$  and airlight $H$. ID is defined in Eq.~\ref{eq: decomposition}. 

\begin{equation}\label{eq: decomposition}
(\theta^*_c, \theta^*_d, \theta^*_m) = \underset{(\theta_c, \theta_d, \theta_m)}{\arg\min} \;\mathcal{L}_{ID} \big(I; f_{\theta_c}, f_{\theta_d}, f_{\theta_m} \big).
\end{equation}
Here, $\mathcal{L}_{ID}$ denotes the image decomposition loss. $\theta_c$ is the parameter of image content layer, $\theta_d$ is the parameter of distortion layer, and $\theta_m$ is the parameter of mask $M$.  $f_{\theta_c}$, $f_{\theta_d}$, and $f_{\theta_m}$ are the instances of encoder-decoder network. $z_c$, $z_d$, and $z_m$ denote the inputs for the networks. Formally, Eq.~\ref{eq: decomposition} models the joint optimization to compute $I^{cfe} = f_{\theta^*_c}(z_c)$, $D = f_{\theta^*_d}(z_d)$, and $M = f_{\theta^*_m}(z_m)$. We have shown $\mathcal{L}_{ID}$ in Eq.~\ref{eq: ldecom}.
\begin{equation}\label{eq: ldecom}
\begin{split}
\mathcal{L}_{ID} \big(I; f_{\theta_c}, f_{\theta_d}, f_{\theta_m} \big) = \Big\| &\big( f_{\theta_m}(z_m) \odot f_{\theta_c}(z_c) + \\ & (1 -  f_{\theta_m}(z_m)) \odot f_{\theta_d} (z_d)\big) - I \Big\|.
\end{split}
\end{equation}
Here, Eq.~\ref{eq: ldecom} shows that the layer separation is achieved by composing image $I$ from image features $I^{cfe} = f_{\theta^*_c}(z_c)$ and corruption layer $D = f_{\theta^*_d}(z_d)$, and then minimizing pixel-wise differences. We will discuss the image decomposition for image dehazing task in Sec.~\ref{sec: dehazing}. 

\begin{figure*}[!h] \centering 
\begin{subfigure}[b]{0.145\linewidth}\captionsetup{justification=centering}\begin{center}
\includegraphics[width=\linewidth]{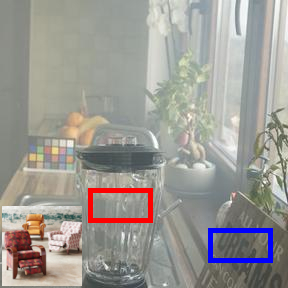} \\
\includegraphics[width=0.445\linewidth, cfbox=red 0.5pt 0.5pt]{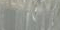}
\includegraphics[width=0.445\linewidth, cfbox=blue 0.5pt 0.5pt]{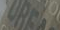}
\end{center} \caption{\footnotesize Content \& Style} \end{subfigure}
\begin{subfigure}[b]{0.145\linewidth}\captionsetup{justification=centering}\begin{center}
\includegraphics[width=\linewidth]{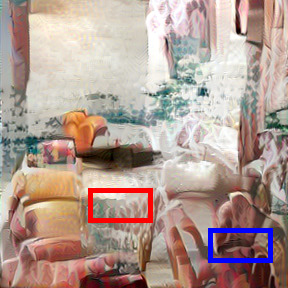} \\
\includegraphics[width=0.445\linewidth, cfbox=red 0.5pt 0.5pt]{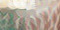}
\includegraphics[width=0.445\linewidth, cfbox=blue 0.5pt 0.5pt]{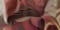}
\end{center} \caption{\footnotesize Neural style  \cite{gatys2016image}} \end{subfigure}
\begin{subfigure}[b]{0.145\linewidth}\captionsetup{justification=centering}\begin{center}
\includegraphics[width=\linewidth]{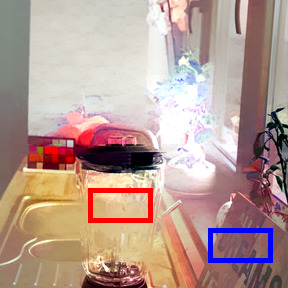} \\
\includegraphics[width=0.445\linewidth, cfbox=red 0.5pt 0.5pt]{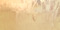}
\includegraphics[width=0.445\linewidth, cfbox=blue 0.5pt 0.5pt]{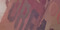}
\end{center} \caption{\footnotesize DPST \cite{luan2017deep} } \end{subfigure}
\begin{subfigure}[b]{0.145\linewidth}\captionsetup{justification=centering}\begin{center}
\includegraphics[width=\linewidth]{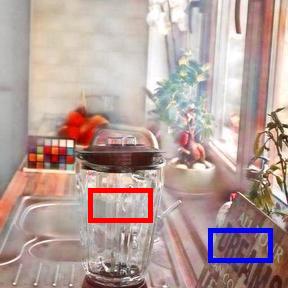} \\
\includegraphics[width=0.445\linewidth, cfbox=red 0.5pt 0.5pt]{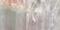}
\includegraphics[width=0.445\linewidth, cfbox=blue 0.5pt 0.5pt]{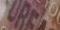}
\end{center} \caption{\footnotesize WCT2 \cite{yoo2019photorealistic}} \end{subfigure}
\begin{subfigure}[b]{0.145\linewidth}\captionsetup{justification=centering}\begin{center}
\includegraphics[width=\linewidth]{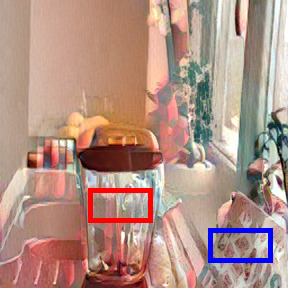} \\
\includegraphics[width=0.445\linewidth, cfbox=red 0.5pt 0.5pt]{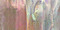}
\includegraphics[width=0.445\linewidth, cfbox=blue 0.5pt 0.5pt]{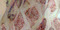}
\end{center} \caption{\footnotesize STROTSS \cite{kolkin2019style}} \end{subfigure}
\begin{subfigure}[b]{0.145\linewidth}\captionsetup{justification=centering}\begin{center}
\includegraphics[width=\linewidth]{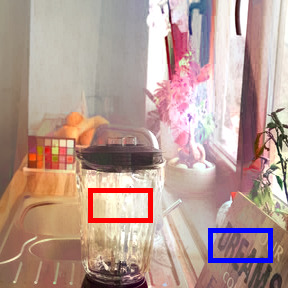} \\
\includegraphics[width=0.445\linewidth, cfbox=red 0.5pt 0.5pt]{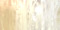}
\includegraphics[width=0.445\linewidth, cfbox=blue 0.5pt 0.5pt]{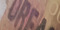}
\end{center} \caption{\footnotesize DILIE (ours)} \end{subfigure}\vspace*{-0.2cm}
\caption{\textbf{Hazy Image Enhancement (indoor).} The figure shows the image feature enhancement of the indoor scene. It could be observed that DILIE (ours) distributed image features with better perceptual quality {\color{blue} (the images are best viewed after zooming)}. }
\label{fig: ihaze}
\end{figure*}


The image decomposition in Eq.~\ref{eq: ldecom} does not consider the context of the input image.  The abstract information of content features represents the context of the image, \textit{i.e.}, objects and their relative positions. The feature extractor VGG19 is denoted by $\phi$. It is used to extract the content features and style features \cite{gatys2016image}. The content features are mostly present at the higher layers of feature extractor $\phi$ denoted by $\phi^C$. The style features are mostly contained at the initial layers denoted by $\phi^S$. The contextual content loss $\mathcal{L}_{CL}$ is defined between the content features of $I$ and $I^{cfe}=f_{\theta_c}(z_c)$ as given in Eq.~\ref{eq: context}. 

\begin{equation}\label{eq: context}
\mathcal{L}_{CL}\big(I, \phi; f_{\theta_c} \big) = - \log CX \big(\phi^C(f_{\theta_c}(z_c)), \phi^C(I) \big).
\end{equation}
Here, $CX$ denotes the contextual similarity computed by finding for each feature $\phi^C(f_{\theta_c}(z_c))_i$ of the image $I^{cfe}$,  the contextually similar feature $\phi^C(I)_j$ of the corrupted image $I$, and then sum over all the features in $\phi^C(f_{\theta_c}(z_c))$. We call the strategy above as the contextual similarity criterion. The key observation is that high-level content information (image context) is similar in both $I^{cfe}$ and $I$.   $\mathcal{L}_{CL}$ maximizes the contextual similarity between $I^{cfe}$ and $I$ to improve performance\footnote{We have used $\phi^C$=\{conv4\_2\} and $\phi^S$=\{conv1\_2, conv2\_2, conv3\_2\} in our experiments. }.

\subsubsection{Image Reconstruction} \label{ssec: IR}
Image reconstruction model (IR) uses encoder-decoder $f$ to reconstruct the desired image. IR is described in Eq.~\ref{eq: reconstruction}.
\begin{equation}\label{eq: reconstruction}
\begin{split}
\theta^* = \;&\underset{\theta}{\arg\min} \;\mathcal{L}_{IR}(I; f_{\theta_r}), \\ & \text{ where} \; \mathcal{L}_{IR}(I; f_{\theta_r}) = \big\| f_{\theta}(z_r) - \mathcal{T}(I) \big\|. 
\end{split}
\end{equation}
Here, $\mathcal{L}_{IR}$ is the reconstruction loss, $\theta_r$ is the network parameters, $z_r$ is the network input, and $\mathcal{T}$ is the image transformation function. The output of CFE in image reconstruction is $I^{cfe} = f_{\theta_r^*}(z_r)$. The function $\mathcal{T}$ varies based on the application under consideration. For example, $\mathcal{T}$ is an identity function for denoising and $\mathcal{T}$ is a downsampling function for super-resolution \cite{ulyanov2018deep}. IR model in Eq.~\ref{eq: reconstruction} performs content feature enhancement by incorporating the application-specific encoder-decoder architectures for $f$. The network architecture is observed to provide an implicit image prior for restoration \cite{ulyanov2018deep}.



\subsection{Style Feature Enhancement} \label{ssec: SFE}
We described that CFE enhances the content features of $I$. SFE aims to improve style features and output the enhanced image $I^*$ given the CFE output $I^{cfe}$. SFE  transfer the style features to $I^{cfe}$ using style image $S$. We define SFE in Eq.~\ref{eq: sfe}. 

\begin{equation}\label{eq: sfe}
I^* = \text{SFE}(I^{cfe}, S, f, \phi, \beta).
\end{equation}
Here, $I^*$ is the enhanced image and $S$ is the reference style image. $\beta$ represents the type of feature enhancement, \textit{i.e.}, photo-realistic ($\beta=1$) or painting style artistic ($\beta=2$). 

The style features enhacement is performed using the content loss $\mathcal{L}_\mathcal{C}$ and style loss $\mathcal{L}_\mathcal{S}$. The content loss $\mathcal{L}_\mathcal{C}$ is defined between the content feature representations $\phi^C(I^{cfe})$ extracted from $I^{cfe}$ and the content feature representations $\phi^C(I^{*})$ extracted from $I^*$. The content loss is given by $\mathcal{L}_\mathcal{C} \;=\; \mathcal{L}\big(\phi^C(I^{cfe}), \phi^C(I^{*})\big)$. The style loss $\mathcal{L}_S$ is computed between the style feature representations $\phi^S(S)$ extracted from $S$ and the style feature representation $\phi^S(I^{*})$ of $I^{*}$. Formally, $\mathcal{L}_S \;=\; \mathcal{L}\big( \phi^S(S), \phi^S(I^{*}) \big)$. We provide the detailed description of $\mathcal{L}_\mathcal{C}$ and $\mathcal{L}_\mathcal{S}$ in the supplementary material. 

SFE could be considered as photo-realistic or artistic features enhancement. The photo-realistic feature enhancement (PE) is aimed to minimize the distortion of object boundaries and preserve photo-realism using loss $\mathcal{L}_{PE}$. In contrast, the artistic feature enhancement (AE) allows small deformations to achieve an artistic look using loss $\mathcal{L}_{AE}$.

\begin{figure*}[!h] \centering
\begin{subfigure}[b]{0.145\linewidth}\captionsetup{justification=centering}\begin{center}
\includegraphics[width=\linewidth]{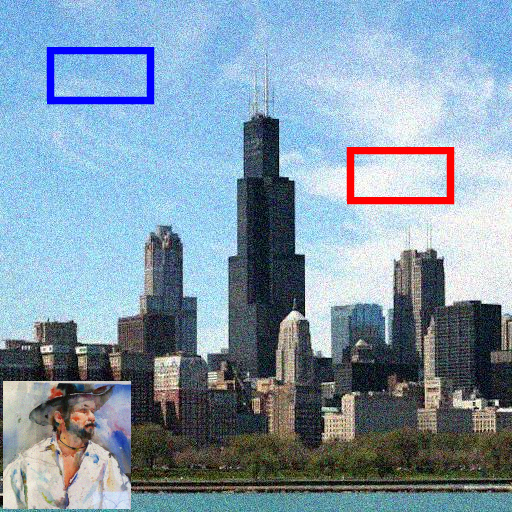} \\
\includegraphics[width=0.44\linewidth, cfbox=red 0.5pt 0.5pt]{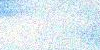}
\includegraphics[width=0.44\linewidth, cfbox=blue 0.5pt 0.5pt]{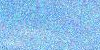}
\end{center} \caption{\footnotesize Content \& Style} \end{subfigure}
\begin{subfigure}[b]{0.145\linewidth}\captionsetup{justification=centering}\begin{center}
\includegraphics[width=\linewidth]{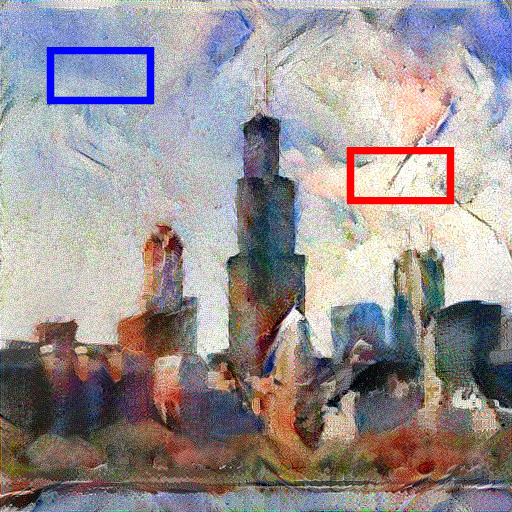} \\
\includegraphics[width=0.44\linewidth, cfbox=red 0.5pt 0.5pt]{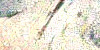}
\includegraphics[width=0.44\linewidth, cfbox=blue 0.5pt 0.5pt]{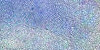}
\end{center} \caption{\footnotesize Neural style \cite{gatys2016image}} \end{subfigure}
\begin{subfigure}[b]{0.145\linewidth}\captionsetup{justification=centering}\begin{center}
\includegraphics[width=\linewidth]{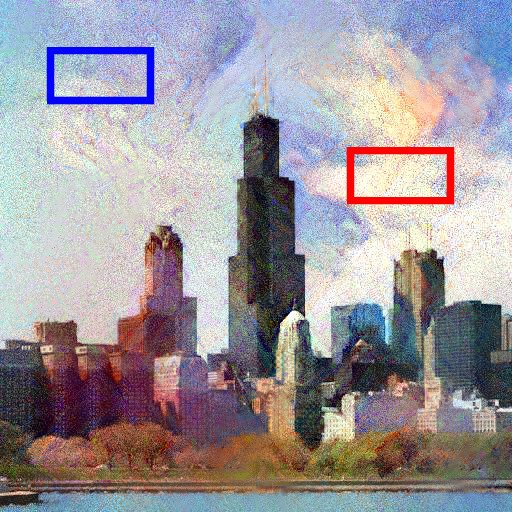} \\
\includegraphics[width=0.44\linewidth, cfbox=red 0.5pt 0.5pt]{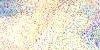}
\includegraphics[width=0.44\linewidth, cfbox=blue 0.5pt 0.5pt]{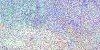}
\end{center} \caption{\footnotesize DPST \cite{luan2017deep}} \end{subfigure}
\begin{subfigure}[b]{0.145\linewidth}\captionsetup{justification=centering}\begin{center}
\includegraphics[width=\linewidth]{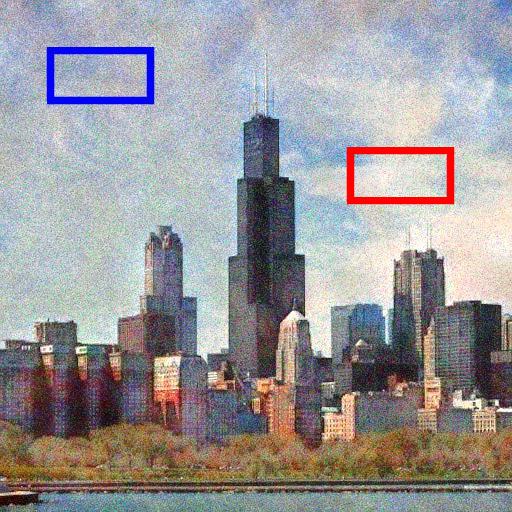} \\
\includegraphics[width=0.44\linewidth, cfbox=red 0.5pt 0.5pt]{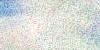}
\includegraphics[width=0.44\linewidth, cfbox=blue 0.5pt 0.5pt]{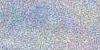}
\end{center} \caption{\footnotesize WCT2 \cite{yoo2019photorealistic}} \end{subfigure}
\begin{subfigure}[b]{0.145\linewidth}\captionsetup{justification=centering}\begin{center}
\includegraphics[width=\linewidth]{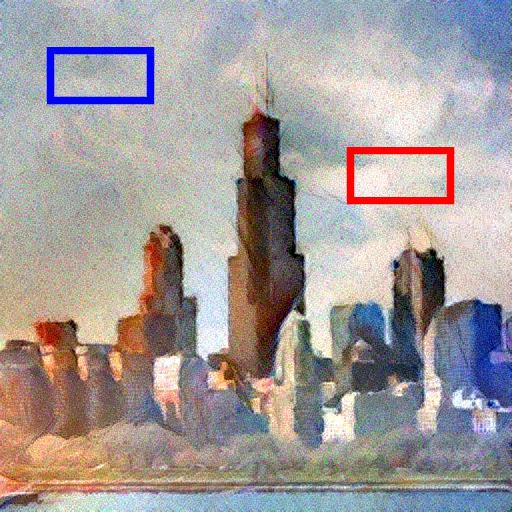} \\
\includegraphics[width=0.44\linewidth, cfbox=red 0.5pt 0.5pt]{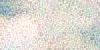}
\includegraphics[width=0.44\linewidth, cfbox=blue 0.5pt 0.5pt]{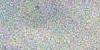}
\end{center} \caption{\footnotesize STROTSS \cite{kolkin2019style}} \end{subfigure}
\begin{subfigure}[b]{0.145\linewidth}\captionsetup{justification=centering}\begin{center}
\includegraphics[width=\linewidth]{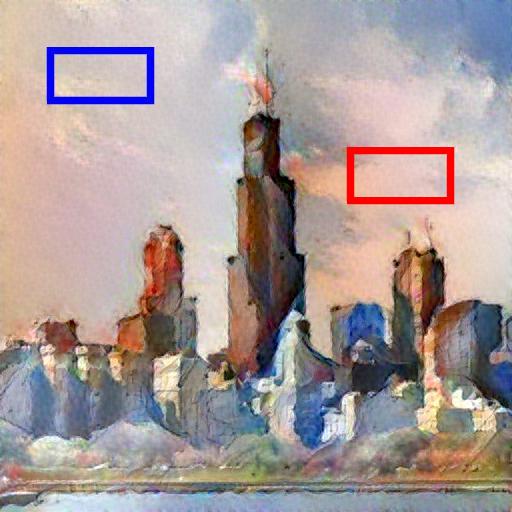} \\
\includegraphics[width=0.44\linewidth, cfbox=red 0.5pt 0.5pt]{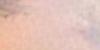}
\includegraphics[width=0.44\linewidth, cfbox=blue 0.5pt 0.5pt]{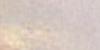}
\end{center} \caption{\footnotesize DILIE (ours) } \end{subfigure}
\caption{\textbf{Noisy image enhancement.} The figure shows that DILIE outputs images with better perceptual quality (see the cropped images). The style images are artistic images and content images contain noise with strength $\sigma=0.25$.}
\label{fig: denoising}
\end{figure*}%

\subsubsection{Photo-realistic Feature Enhancement} \label{ssec: PE}
The photo-realism characterization in the image is an unsolved problem \cite{luan2017deep}. The enhancement of the photo-realistic features is based on the observation that if the input image is photo-realistic, then those features could be retained with an affine loss \cite{luan2017deep}.  The image with lower perceptual errors is observed to be more photo-realistic \cite{prashnani2018pieapp}. The degree of photo-realism in the output $I^*$ is measured by the perceptual error score PieAPP \cite{prashnani2018pieapp}.

The total loss for PE is defined as $\mathcal{L}_{PE} = \mathcal{L}_m + \mu \times \mathcal{L}_\mathcal{C} + \kappa \times \mathcal{L}_\mathcal{S}$, where $\mu$ and $\kappa$ are the coefficients for the content loss $\mathcal{L}_\mathcal{C}$ and the style loss $\mathcal{L}_\mathcal{S}$. The affine loss $\mathcal{L}_m$ preserves the object structure while transforming the style features. More specifically, affine loss uses Matting Laplacian $\mathcal{M}_{I^{cfe}}$ of the input $I^{cfe}$  \cite{luan2017deep}, where $\mathcal{M}_{I^{cfe}}$ represents the grayscale matte for the content features. Intuitively, the affine loss function transforms the color distribution of $I^*$ while preserving the object structure. 



\subsubsection{Artistic Feature Enhancement} \label{ssec: AE}
We described that small image feature deformation could be present in the artistic style transfer. Therefore, the strategy is to match the distribution of the style and the content features and do not use the affine loss to reduce deformations in $I^*$. 
  
The total loss for AE is defined as $\mathcal{L}_{AE} = \mu \times \mathcal{L}_\mathcal{C} + \kappa \times \mathcal{L}_\mathcal{S}$, where $\mu$ and $\kappa$ are the coefficients for the content loss $\mathcal{L}_\mathcal{C}$ and the style loss $\mathcal{L}_\mathcal{S}$.  We use relaxed earth mover distance (EMD) to match the image feature distribution \cite{kolkin2019style}. The EMD loss preserves the distance between all the pairs of features extracted from the VGG19 $\phi$ to allow pixel value modification for style features while preserving the structure of the objects.



\begin{table*}[!h]  \centering \setlength\extrarowheight{3pt} 
\caption{\small{The table shows SSIM comparison for dehazing of I-Haze and O-Haze dataset. DILIE outperforms other methods in comparison.}}
\label{table: SSIM}
\resizebox{0.84\linewidth}{!}{
\begin{tabular}{|c|c|c|c|c|c|c|c|c|c|c|}
\hline 
       & \textbf{AODNet \cite{li2017aod}} & \textbf{MSCNN \cite{ren2016single}} & \textbf{DcGAN \cite{li2018single}} & \textbf{GFN \cite{ren2018gated}} & \textbf{GCANet \cite{chen2019gated}} & \textbf{PFFNet \cite{mei2018progressive}} & \textbf{DoubleDIP \cite{gandelsman2019double}} & \textbf{DILIE (ours)} \\ \hline
\textit{I-Haze \cite{I-HAZE_2018}} & 0.732           & 0.755          & 0.733          & 0.751        & 0.719           & 0.740           & 0.691 &         \textbf{0.790}  \\ \hline
\textit{O-Haze \cite{O-HAZE_2018}} & 0.539           & 0.650          & 0.681          & 0.671        & 0.645           & 0.669           & 0.643 &       \textbf{0.705}         \\ \hline
\end{tabular}}
\end{table*}
\begin{table}[!h] \centering  \setlength\extrarowheight{3pt} 
\caption{\small{The table shows that DILIE (ours) performs image enhancement with minimum perceptual error PieAPP  \cite{prashnani2018pieapp}.}}
\label{table: PieAPP_NIMA}
\resizebox{\linewidth}{!}{
\begin{tabular}{|c|c|c|c|c|c|}
\hline
       & \textbf{Neural \cite{gatys2016image}} & \textbf{DPST \cite{luan2017deep}} & \textbf{WCT2 \cite{yoo2019photorealistic}} & \textbf{STROTSS \cite{kolkin2019style}} & \textbf{DILIE} \\ \hline
\textit{I-Haze \cite{I-HAZE_2018}} & 3.80 & 3.33 & 3.52 & 2.91 & \textbf{2.78} \\ \hline
\textit{O-Haze \cite{O-HAZE_2018}} &   3.00 &  2.71 & 2.88  &   2.81 & \textbf{2.55} \\ \hline
\textit{Denoising 100} & 5.00 & 4.98  &  4.53  &  4.82  &  \textbf{4.27} \\ \hline
\end{tabular}}
\end{table}%

\section{Applications} \label{sec: applications}
We perform image enhancement of hazy and noisy images. 

\subsection{Hazy Image Enhancement} \label{sec: dehazing} 
Pictures taken in the hazy weather may lack scene information such as contrast, colors, and object structure. Haze is composed of small particles (e.g., dust) suspended in the gas. We have discussed the pictorial representation for hazy image enhancement in Sec.~\ref{sec: methodology} (Fig.~\ref{fig: model}). The image degradation model for the hazy image is usually formulated using an atmospheric scattering model \cite{yang2018proximal} as shown in Eq.~\ref{eq: hazyModel}.
\begin{equation}\label{eq: hazyModel}
I(p) = \hat{I}(p) \times M(p) + H(p) \times (1 - M(p)).
\end{equation}
Here, $p$ is the pixel location and $I$ is the degraded observation. $\hat{I}$ is the haze-free image and $M$ is the transmission map. Intuitively, the hazy image $I$ could be considered as a haze layer $H$ superimposed on the true scene content $\hat{I}$. 

Image dehazing can be formulated as a layer decomposition problem to separates the hazy image ($I$) into a haze-free image layer ($I^{cfe}$) and a haze layer ($H$), where $I^{cfe}$ is the approximation of haze-free image $\hat{I}$. We have discussed the generalized image decomposition framework in Eq.~\ref{eq: decomposition}. We show its applicability for hazy image enhancement in Eq.~\ref{eq: hazeDecom}. 

\begin{equation}\label{eq: hazeDecom}
(\theta^*_c, \theta^*_h, \theta^*_m) = \underset{(\theta_c, \theta_h, \theta_m)}{\arg\min} \; \mathcal{L}_{ID} \big(I; f_{\theta_c}, f_{\theta_h}, f_{\theta_m} \big) + \mathcal{L}_{CL}\big(I, \phi; f_{\theta_c} \big).
\end{equation}

Here, $\mathcal{L}_{ID}$ is for image decomposition (Eq.~\ref{eq: decomposition}) and $\mathcal{L}_{CL}$ is for preserving image context  (Eq.~\ref{eq: context}). $\theta_h$ represents the parameters for haze layer. The transmission map $M=f_{\theta^*_m}(z_m)$ separates the haze-free image $I^{cfe}= f_{\theta^*_c}(z_c)$ and the atmospheric light $H= f_{\theta^*_h} (z_h)$. The joint framework is aimed to estimate $\hat{I}$ and $H$ preserving their relations. 


\begin{figure*}[!ht]\centering
\begin{subfigure}[b]{0.125\textwidth}\captionsetup{justification=centering} \includegraphics[width=\linewidth]{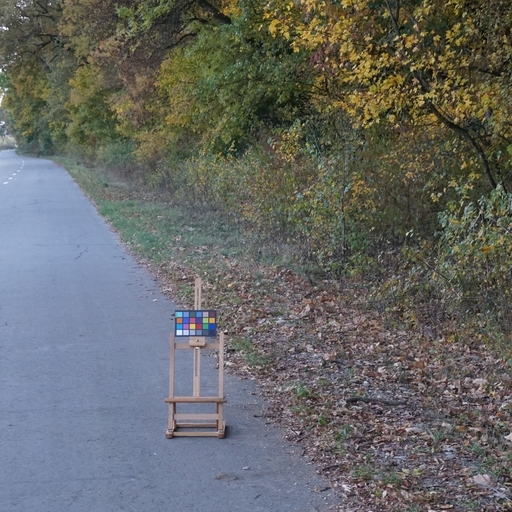} \caption{\footnotesize Haze-free  Image $\hat{I}$} \end{subfigure}
\begin{subfigure}[b]{0.125\textwidth} \captionsetup{justification=centering}\includegraphics[width=\linewidth]{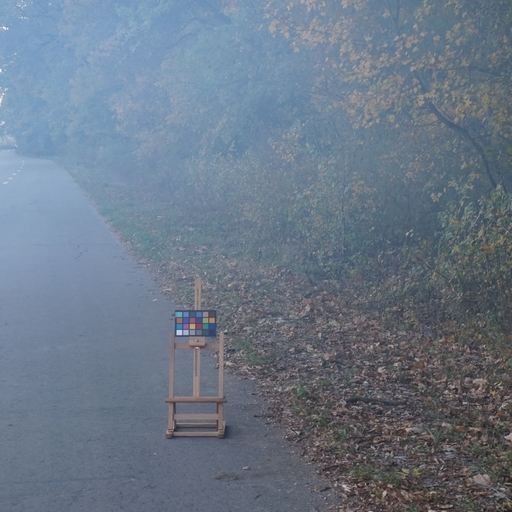} \caption{\footnotesize  Hazy Image \\$I$} \end{subfigure}
\begin{subfigure}[b]{0.125\textwidth} \captionsetup{justification=centering}\includegraphics[width=\linewidth]{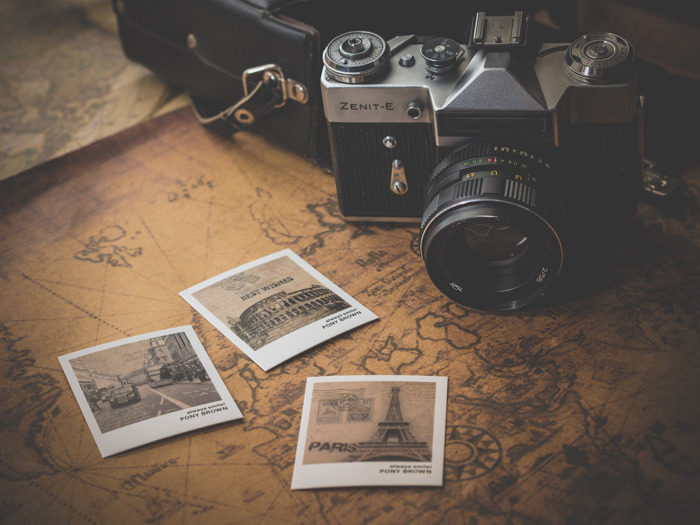} \caption{\footnotesize  Style Image \\$S$} \end{subfigure}
\begin{subfigure}[b]{0.125\textwidth} \captionsetup{justification=centering}\includegraphics[width=\linewidth]{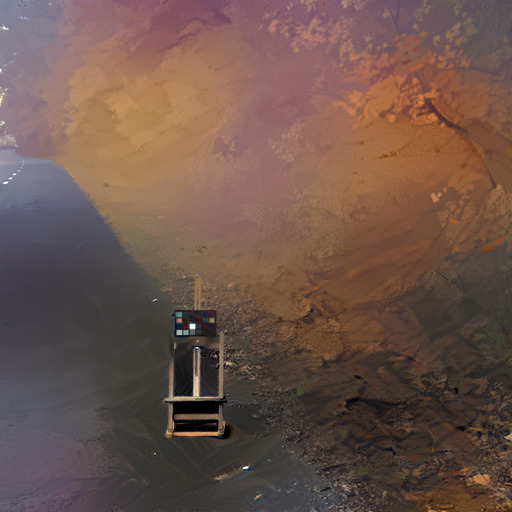} \caption{\footnotesize DPST \cite{luan2017deep} \\ \emph{H}: 0.455} \end{subfigure}
\begin{subfigure}[b]{0.125\textwidth}\captionsetup{justification=centering} \includegraphics[width=\linewidth]{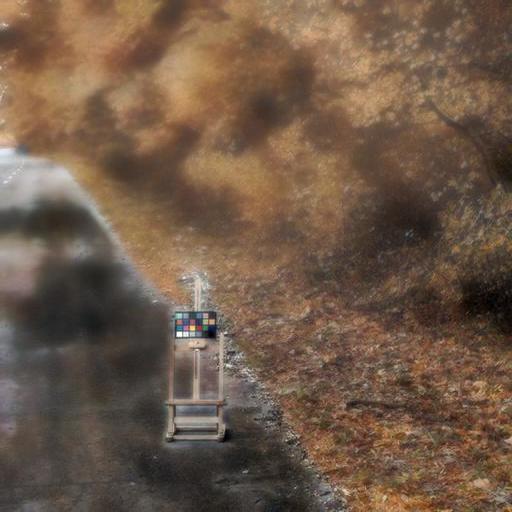} \caption{\footnotesize WCT2 \cite{yoo2019photorealistic} \\ \emph{H}: 0.943} \end{subfigure}
\begin{subfigure}[b]{0.125\textwidth} \captionsetup{justification=centering}\includegraphics[width=\linewidth]{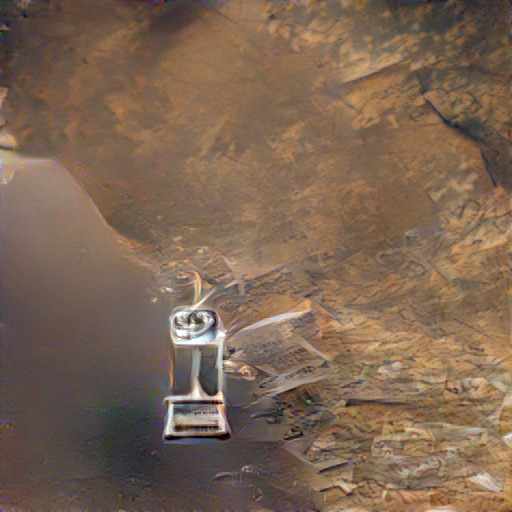} \caption{\footnotesize STROTSS \cite{kolkin2019style} \\ \emph{H}: 0.876} \end{subfigure}
\begin{subfigure}[b]{0.125\textwidth} \captionsetup{justification=centering}\includegraphics[width=\linewidth]{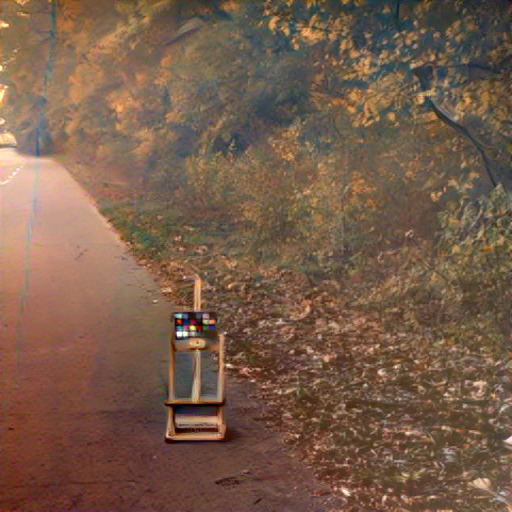} \caption{\footnotesize DILIE (ours) \\ \emph{H}: 0.350} \end{subfigure}
 \caption{\textbf{Ablation Study.} The figure highlights the corruption of image features due to the haze in the enhanced output images. The style features (color information) of the outputs get affected by haze even when the input style image does not contain haze particles. \emph{H} denotes the relative perceptual error due to haze computed using PieAPP \cite{prashnani2018pieapp}. DILIE output image with the minimum perceptual error. It could also be observed visually that DILIE output has minimum effect from the haze.}
\label{fig: ablation}
\end{figure*}
 
The main goal of Eq.~\ref{eq: hazeDecom} is to separate image features and haze features based on the semantics. The characteristics of haze particles in $I$ are similar. Therefore, they accumulate into haze layer $H$. Similarly, the image features of $I$ have similar characteristics and get separated into the haze-free image layer $I^{cfe}$. We have discussed contextual content loss $\mathcal{L}_{CL}$ given in Eq.~\ref{eq: context} matches the contextual similarity between features. $\mathcal{L}_{CL}$ improves the performance of the layer decomposition framework. 

Fig.~\ref{fig: ohaze} shows the image enhancement of outdoor images and Fig.~\ref{fig: ihaze} shows the enhancement of the indoor scenes. The outdoor scenes mostly contain clouds and trees and the indoor images mostly contain objects present in the household. The image dehazing removes the haze from the input image and hazy image enhancement improves the quality of image features.

Table~\ref{table: SSIM}  shows that DILIE achieves a good Structural Similarity Index (SSIM) for image dehazing. Table~\ref{table: PieAPP_NIMA} shows that DILIE output images with better perceptual quality for hazy image enhancement\footnote{We used implementation of Neural style provided in \cite{Smith2016}, Tensorflow implementation of DPS given in \cite{YangPhotoStyle2017}, contextual loss implementation in \cite{roimehrez2018}, STROTSS implementation in \cite{nkolkin13STROTSS}, and WCT2 implementation in \cite{clovaaiWCT2}.  We have provided more visual comparisons and implementation details of our method in the supplementary material.}. It is interesting to observe that the generalisability of DILIE (ours) allows good performance for both content feature enhancement (image dehazing) and style feature enhancement (hazy image enhancement).

Fig.~\ref{fig: ablation} shows that if the input image contains haze particles, then the haze information gets incorporated into the output even when $S$ does not include haze information. Ideally, the output should contain the content features from  $I$ and style features from $S$. The image enhancement of hazy images highlight that preserving a perceptually good balance between style and content features is very challenging. Our CFE module removes haze features so that the final output $I^*$ has less influence due to bad weather conditions. 


\subsection{Noisy Image Enhancement}
Denoising aims to recover a clean image from a noisy observation. The image degradation model for the noisy image is given as $I = \hat{I} + \epsilon$. Here, $I$ the noisy image, $\hat{I}$ is the clean content image, and $\epsilon$ is the additive noise. 

Image denoising is formulated as image reconstruction, where an encoder-decoder $f$ reconstructs the clear image $I^{cfe}$ from the noisy observation $I$. The network $f$ provides a high impedance to noise and allows image features \cite{ulyanov2018deep}. We have discussed the generalized framework for image reconstruction using transformation $\mathcal{T}$ in Eq.~\ref{eq: reconstruction}. Image denoising is performed by taking $\mathcal{T}$ to be identity function as given in Eq.~\ref{eq: denoising}. 
\begin{equation}\label{eq: denoising}
I^{cfe}=f_\theta (z), \; \text{where}, \;\theta^* =  \underset{\theta }{\arg\min}  \| \; f_\theta (z) - I \; \|.
\end{equation}
Here, the restored image $I^{cfe}=f_\theta (z)$ is the approximation of $\hat{I}$. The reconstruction loss given in Eq.~\ref{eq: denoising} is iteratively minimized, and early stopping is used to get the best possible outcome before the network over-learn the noisy features. 

We make noisy image enhancement more challenging by using the style and the content images containing noise with the strength $\sigma=0.25$. We show the output images in Fig.~\ref{fig: denoising}. It could be observed that DILIE gets a better distribution of features with better clarity (see cropped images). We have shown a quantitative comparison in Table~\ref{table: PieAPP_NIMA}. It can be observed that DILIE outperforms other methods in comparison.

\section{Ablation Study}
Fig.~\ref{fig: ablation} illustrates that DILIE output images with less environmental noise. The quantitative comparison for haze corruption is described as follows. Consider the hazy image $I$, haze-free image $\hat{I}$, and the style image $S$ (Fig.~\ref{fig: ablation}). The difference of image features between $I$ and $\hat{I}$ is due to the haze. Let ST($y$, $z$) denote the style transfer of content $y$ using style $z$. Fig.~\ref{fig: ablation} shows that when performing ST between $I$ and $S$, the output image is observed to have haze corruption even when $S$ does not have haze information. The goal is to minimize haze corruption.

To quantify haze corruption, let \emph{E($w$, $x$)} denote the perceptual error \cite{prashnani2018pieapp} between image $w$ and image $x$. The relative error \emph{H} = $\|$\emph{E}\big($\hat{I}$, ST($\hat{I}$, $S$)\big) $-$ \emph{E} \big($\hat{I}$, ST($I$, $S$)\big)$\|$ with reference to haze-free image $\hat{I}$ measures the deformations caused by haze in ST($I$, $S$) by comparing ST output of the clean image $\hat{I}$ and the corrupted image $I$ using perceptual error PieAPP \cite{prashnani2018pieapp}. 

Fig.~\ref{fig: ablation} shows that DILIE output image with minimum perceptual error \emph{H}. It could also be observed visually that in WCT2 \cite{yoo2019photorealistic} output contains haze corruption. DPST \cite{luan2017deep} and STROTSS \cite{kolkin2019style} outputs also have haze effects when looking carefully. DILIE has the minimum haze effect \footnote{We discuss the ablation study more in the supplementary material.}.

\section{Conclusion}
DILIE is a deep internal learning approach for image enhancement.  It is a generic framework for image restoration and image style transfer tasks for content feature enhancement (CFE) and style feature enhancement (SFE) models. The contextual content loss for image decomposition improvised the performance of the image dehazing task. The interesting challenge here is that the degraded input image corrupts both style and content features. CFE and SFE together lead to output images with a low perceptual error and good structure similarity. As future work, we propose to explore image enhancement for other image degradation models such as under-water scenes and snowfall.

{\small
\bibliographystyle{ieee_fullname}
\bibliography{egbib}
}

\end{document}